\pdfoutput=1

\documentclass[11pt]{article}

\usepackage{naacl2021}

\usepackage{times}
\usepackage{latexsym}

\usepackage[T1]{fontenc}

\usepackage[utf8]{inputenc}

\usepackage{microtype}
\usepackage{microtype}

\usepackage{xcolor}
\usepackage{colortbl}
\usepackage{hhline}
\usepackage{amsmath}
\usepackage{amssymb}
\usepackage{multirow}
\usepackage{booktabs}
\usepackage{graphicx}
\usepackage{enumitem}
\usepackage{tikz}
\usepackage{booktabs}
\usepackage{xspace}
\usepackage{comment}

\usepackage[ruled,noline,linesnumbered]{algorithm2e}
\usepackage{algorithm2e}
\usepackage{algorithmic}

\usepackage{xspace,mfirstuc,tabulary}

\newif\iftaclinstructions
\taclinstructionsfalse 
\iftaclinstructions

\newcommand{\instr}
\fi

\newcommand{\method}{\frenchspacing\textsc{NeuroLogic Decoding}\xspace}
\newcommand{\methodshort}{\frenchspacing\textsc{NeuroLogic}\xspace}
\newcommand{\commongen}{\frenchspacing\textsc{CommonGen}\xspace}

\definecolor{amber(sae/ece)}{rgb}{1.0, 0.49, 0.0}
\definecolor{cobalt}{rgb}{0.0, 0.28, 0.67}

\title{
\textsc{NeuroLogic Decoding:} \\
(Un)supervised Neural Text Generation with Predicate Logic Constraints
}

\author{Ximing Lu\textsuperscript{$\dagger\ddagger$} \hspace{.3cm}   Peter West\textsuperscript{$\dagger\ddagger$}  \hspace{.3cm}
Rowan Zellers\textsuperscript{$\dagger\ddagger$}
\\ 
\textbf{ 
 Ronan Le Bras\textsuperscript{$\ddagger$}
 \hspace{.3cm}
Chandra Bhagavatula\textsuperscript{$\ddagger$}
\hspace{.3cm} Yejin Choi\textsuperscript{$\dagger\ddagger$} }\\
  \textsuperscript{$\dagger$}Paul G. Allen School of Computer Science \& Engineering, University of Washington\\
  \textsuperscript{$\ddagger$}Allen Institute for Artificial Intelligence\\
  \texttt{\{lux32, pawest, rowanz, yejin\}@cs.washington.edu} \\
  \texttt{\{ronanlb, chandrab\}@allenai.org}
  }

\date{}

\begin{document}
\maketitle
\begin{abstract}

Conditional text generation often requires lexical constraints, i.e., which words should or shouldn't be included in the output text. 
While the dominant recipe for conditional text generation has been large-scale pretrained language models that are finetuned on the task-specific training data, such models do not learn to follow the underlying constraints reliably, even when supervised with large amounts of task-specific examples. 

We propose \textsc{NeuroLogic Decoding}, a simple yet effective algorithm that enables neural language models -- supervised or not -- to generate fluent text while satisfying complex lexical constraints. Our approach is powerful yet efficient. It handles any set of lexical constraints that is expressible under predicate logic, while its asymptotic runtime is equivalent to conventional beam search. 

Empirical results on four benchmarks show that \textsc{NeuroLogic Decoding} outperforms previous approaches, including algorithms that handle a subset of our constraints. 
Moreover, we find that unsupervised models with \textsc{NeuroLogic Decoding} often outperform supervised models with conventional decoding, even when the latter is based on considerably larger networks. Our results suggest the limit of large-scale neural networks for fine-grained controllable generation and the promise of inference-time algorithms. 
\end{abstract}

\section{Introduction}
\begin{figure}[t!]
\centering
    \includegraphics[width=0.95\linewidth, height=12cm]{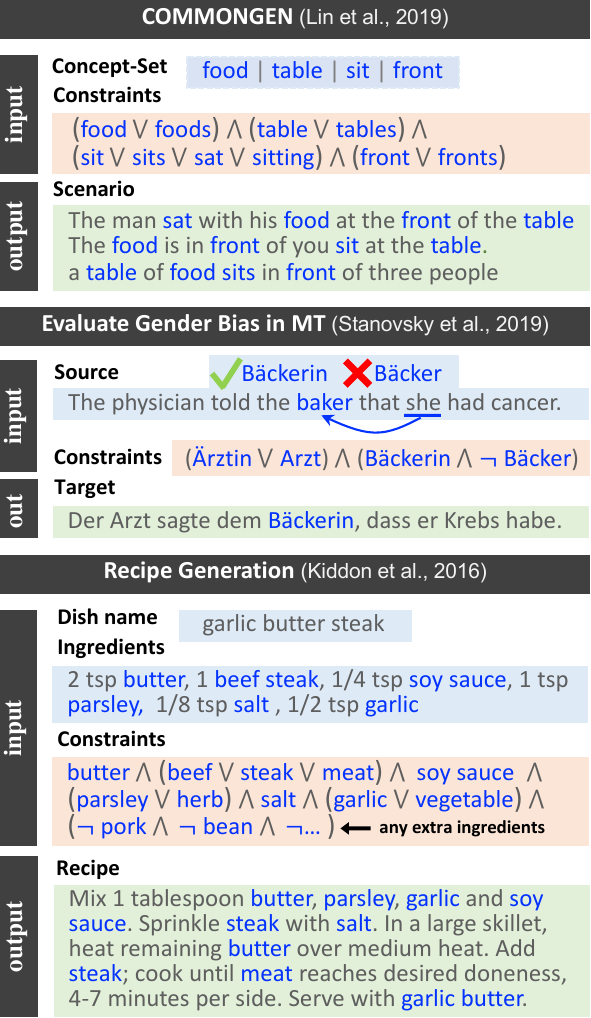}
\caption{Overview of several constrained generation tasks. For instance, generating a short description from a set of concepts \citep[\commongen;][]{lin2019commongen} requires using each of those words at least once; this can be expressed as a logical expression (here, `($\textrm{food} \lor \textrm{foods})\land\ldots$'). Our proposed \method~handles all predicate logic constraints efficiently, yet with the same asymptotic runtime as beam search.}
 \label{tab:fig1}
\end{figure}

Text generation applications often need to incorporate semantic constraints, i.e., what words \emph{should} and \emph{shouldn't} appear in the output generation. 
%
Consider the task of generating a recipe from a set of ingredients \cite{kiddon-etal-2016-globally}, such as `garlic,' `steak', and `soy sauce' (Figure~\ref{tab:fig1}). A generated recipe should cover all of those ingredients, without hallucinating new ones (such as `pork' or `beans'). This restriction, like others in Figure \ref{tab:fig1} for other applications, can be modeled by a set of lexical constraints expressed as a predicate logic formula.

The dominant paradigm today for performing such \emph{constrained generation} is to start with a pretrained language model, and then finetune it on a dataset of task-specific examples. However, pretrained language models struggle at learning to follow these constraints, even when the finetuning dataset is large. For example, for the aforementioned recipe generation task, a GPT2 model finetuned on hundreds of thousands of recipes still hallucinates extra ingredients.  
In stark contrast, humans need to see only a few examples (or even none) to generate the desired output satisfying all the logical constraints, e.g., writing a recipe that 
mentions each ingredient (butter, steak, etc.) without using new ones.

We hypothesize that this mismatch is due to a fundamental under-specification of finetuning. If we finetune one of today's state-of-the-art language models on a dataset, the likelihood of it generating sequences from the same distribution should increase. Yet there is no guarantee that this improvement in likelihood will come from improvements on the fundamental task of constrained generation, as opposed to picking up on dataset-specific patterns such as language style. 
In fact, we present analysis suggesting that `worst-case' learning behavior is common in practice: when we increase the finetuning data fed to GPT2 by an order of magnitude, constraint-satisfaction with standard beam search shows only modest improvement.

To address this issue, we propose \method, which effectively enforces the satisfaction of given lexical constraints by controlling the decoding stage of sequence generation. These constraints can be any predicate logic formula, which crucially includes both positive constraints (the word `butter' must be generated somewhere) and negative constraints (`bean' cannot be generated). These simpler constraints can then be combined through logical connectives to handle more complex requirements such as inflection or synonyms (`beef' \emph{or} `steak' both satisfy the constraint of referring to the steak). While beam search aims to maximize the likelihood of the generated sequence, our method searches for optimal output sequences among the strings that also satisfy the given constraints. It does so efficiently: we convert the hard logic constraints into a soft penalty term in the decoding objective, and use a beam-based search to find approximately-optimal solutions; constraint states are tracked to reuse computation. \method thus effectively and efficiently controls text generation without requiring any modification of the model structure or training pipeline.


We evaluate our method on four different text generation tasks: generative commonsense reasoning (C{\scriptsize OMMON}G{\scriptsize EN}; \citealp{lin2019commongen}), recipe generation \cite{kiddon-etal-2016-globally}, data-grounded dialogue response generation \cite{wen-etal-2015-semantically}, and reducing gender bias in machine translation \cite{stanovsky-etal-2019-evaluating}. 
Empirical results demonstrate that \method ensures the satisfaction of given constraints while maintaining high generation quality, in turn leading to new SOTA results in both the supervised and zero-shot setting.

\makeatletter
\newcommand{\stx}[1]{\mathpalette\giusti@stx{#1}}
\newcommand{\giusti@stx}[2]{%
  \mbox{%
    \medmuskip=\thinmuskip
    \thickmuskip=\thinmuskip
    $\m@th#1#2$%
  }%
}
\makeatother
\section{Method}
In this section, we first rigorously define predicate logic constraint, and then present in detail the \method algorithm.

\subsection{Predicate Logic Constraint}
Let us define a predicate $D(\textbf{a}, \textbf{y})$ to be a boolean function indicating the occurrence of key phrase $\textbf{a}$ in a sequence $\textbf{y}$, where \textbf{a} can be either unigram or multi-gram. $D(\textbf{a}, \textbf{y})$ will be true iff $\textbf{a}$ occurs in \textbf{y}.
\begin{align*}
    &D(\textbf{a}, \textbf{y}) \equiv \exists \: i, \> \textbf{y}_{i:i+|\textbf{a}|} = \textbf{a}
\end{align*}
\methodshort accepts lexical constraints in Conjunctive Normal Form (CNF):
\[\underbrace{\big(D_1 \lor D_2 \cdots \lor D_i \big)}_{C_1}\land \cdots \land \underbrace{\big(D_k \lor D_{k+1} \cdots \lor D_n \big)}_{C_m} \]
where each $D_i$ represents a single positive or negative constraint,  $D(\textbf{a}_\textbf{i}, \textbf{y})$ or $\neg D(\textbf{a}_\textbf{i}, \textbf{y})$, restricting whether key phrase $\textbf{a}_\textbf{i}$ should be strictly included or omitted in \textbf{y}, respectively. Any propositional logical formula can be converted to CNF, and thus handled by \methodshort. Notationally, we will refer to each individual constraint $D_i$ as a \emph{literal}, and the disjunction of literals as a \emph{clause}, denoted as $C_j$, with $L$ being the total number of clauses. Our method seeks optimal sequences in which all clauses are satisfied: 

\vspace*{-9mm}
\begin{equation}
    \hat{\textbf{y}} {=} \arg\max_{\textbf{y} \in \mathcal{Y}}P_\theta(\textbf{y}|\textbf{x}) \hspace{1em}\textrm{where}\hspace{1em} \sum_{i=1}^{L} C_i{=}L
 \end{equation}
Past work on constrained optimization introduces penalties \cite{Fiacco1976SensitivityAF} to approximate the constrained optimization problem with an unconstrained problem. Specifically, by adding a high-cost penalty term for violated constraints:

\vspace*{-5mm}
\begin{align}
    \hat{\textbf{y}} = &\arg\max_{\textbf{y} \in \mathcal{Y}}P_\theta(\textbf{y}|\textbf{x}) - \lambda' \sum_{i=1}^{L} (1-C_i)
 \end{align}
Intuitively, this objective balances sequence likelihood (term 1) and constraint satisfaction (term 2). The aim is to find sequences that do well at both dimensions. While exhaustive search is intractable, we use a beam-based search to find approximately-optimal solutions for this objective. 
\begin{figure}[t]
 \centering
     \includegraphics[width=0.99\linewidth]{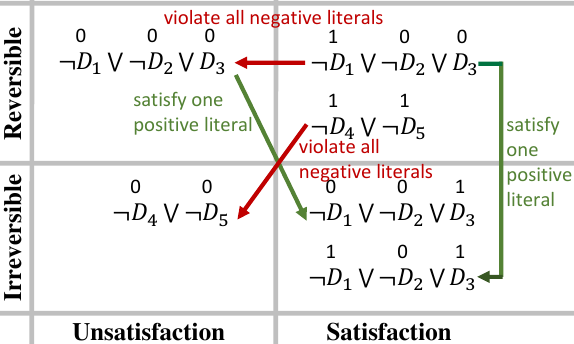}
      \caption{Clause states and possible transitions. $D_i$ and $\neg D_i$ denote positive and negative literal respectively.} 
      \label{fig:clausefigure}
 \end{figure} 
\begin{figure*}[t]
 \centering
     \includegraphics[width=1\linewidth]{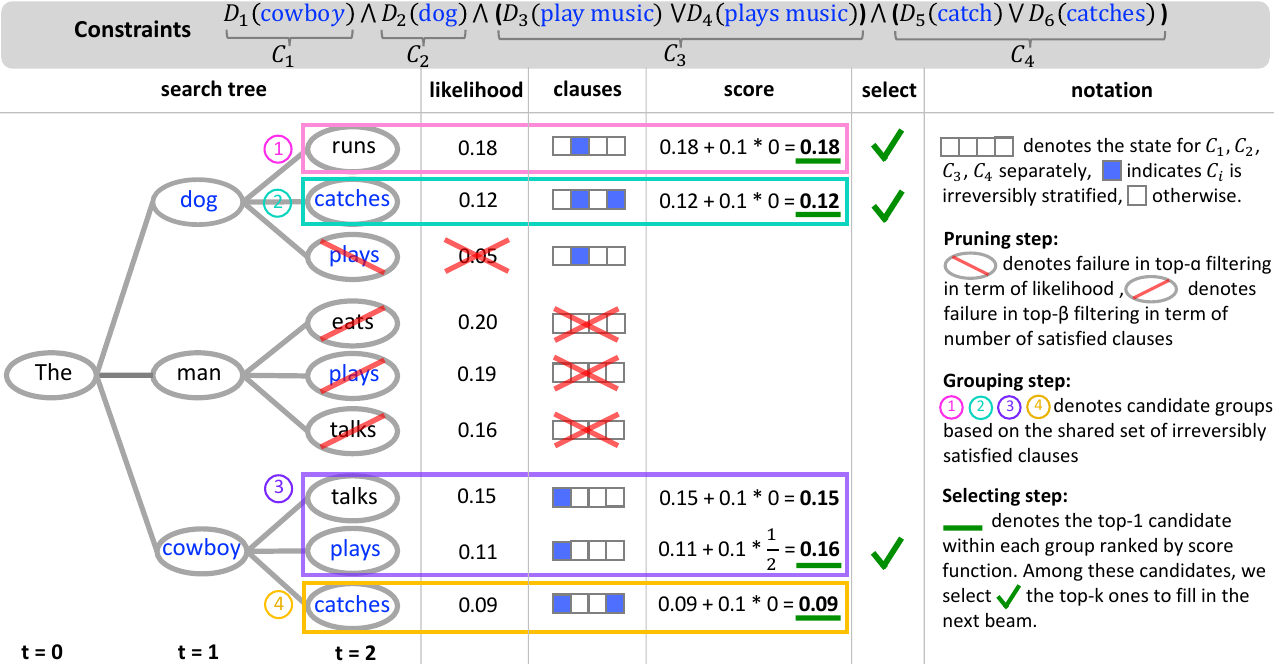}
      \caption{Illustration of the \methodshort decoding procedure. In this example, $k=3$, $\alpha=8$, $\beta=2$, $\lambda=0.1$
      } 
      \label{fig:alg}
 \end{figure*} 
\subsection{Constraint States} \label{section: state}
When considering whether a generation hypothesis satisfies some clause $C_i$ during generation, there are fundamentally 4 possible states (as in figure \ref{fig:clausefigure})
\begin{enumerate}[topsep=0pt, itemsep=0pt, partopsep=0pt,parsep=0pt]
\item[S1] \textbf{reversible unsatisfaction}: If an unsatisfied clause $C_i$ contains at least one positive literal, $C_i$ could be satisfied in the future by fulfilling one of its positive literal(s).
\item[S2] \textbf{irreversible unsatisfaction}: If an unsatisfied clause $C_i$ contains negative literal(s) only, $C_i$ will maintain unsatisfied in the future since the violation of negative literals could not be overturned.
\item[S3] \textbf{reversible satisfaction}: If all satisfied literal(s) in a satisfied clause $C_i$ are negative literal(s), $C_i$ could switch back to unsatisfied in the future by violating all of its satisfied negative literal(s).
\item[S4] \textbf{irreversible satisfaction}: If satisfied literal(s) in a satisfied clause $C_i$ contains at least one positive literal, $C_i$ will maintain satisfied in the future since the fulfilment of positive literals is irreversible.
\end{enumerate}

\noindent 
To track 
the states of literals and clauses efficiently, 
we maintain two prefix tries. The first trie, $\mathcal{T}^{+}$, tracks \emph{unsatisfied positive} literals from all clauses in states S1 and S3, while the other trie, $\mathcal{T}^{-}$, tracks  \emph{satisfied negative} literals from all clauses in state S3. 
We do not track anything from clauses in state S2 or S4, as those are already irreversible. 

If a positive literal is satisfied, its clause in state S1 or S3 is henceforth irreversibly  satisfied (state S4), thus we remove all literals of that clause from both tries and stop tracking. If a negative literal in state S3 is violated, we remote it from the trie $\mathcal{T}^{-}$. Once all negative literals of a clause in state S3 has been removed, the clause switches back to unsatisfied (state S1 or S2). If it has unsatisfied positive literal(s) in the trie $\mathcal{T}^{+}$, it becomes reversibly unsatisfied (state S1); otherwise it shall stay irreversibly unsatisfied (state S2).

\subsection{Algorithm} \label{section: algo}

Since exhaustive search to optimize the CNF constraints is intractable, \methodshort uses a beam-based search to approximate.
The high-level intuition is that at each time step, \methodshort selects generation hypotheses in consideration of both the objective function and the diversity of the partially satisfied constraints. 
We achieve such by 3 steps: \emph{pruning}, \emph{grouping}, and \emph{selecting} (illustrated in figure \ref{fig:alg}, and detailed below).

At each time step, the decoding model generates a distribution over all vocabulary $V$ for $k$ hypotheses in the current beam, resulting in a candidate score matrix of size $k \times |V|$. Along with generating score matrix, we produce a constraint state for each of the $k \times |V|$ new candidates $h$, based on the next token considered.

\textbf{Pruning step}: We first discard any $h$ with irreversible unsatisfied clauses (state S2) to focus only on candidates that might satisfy all constraints. Then, we filter candidates $h$ to those in the top-tier of both satisfied constraints and sequence likelihood. Specifically, we drop any candidates not in the top-$\alpha$ in terms of likelihood $P_\theta(\textbf{y}_t|\textbf{y}_{<t})$, and not in the top-$\beta$ in terms of number of satisfied clauses $\sum_{i=1}^{L} C_i$. These are adjustable parameters, corresponding to maximum tolerance to sequence fluency and constraint satisfaction. 

\textbf{Grouping step}: Next, we select the beam from the pruned candidates. Naively selecting $k$ best candidates with respect to the objective function would not work well, since such greedy selection would bias toward sequences with high likelihood and easy-to-satisfy clauses at early timestep, which can lead to struggling with remaining hard-to-satisfy clauses later on. Therefore, the key intuition is to consider diverse partial solutions early on with respect to the set of irreversibly satisfied clauses, i.e., $\{C_i \> | \> C_i \in \textnormal{state S4}\}$. We group candidates based on this set and select (in the next step) the best ones from each group to fill the beam. 

\textbf{Selecting step}: To select best ones from each group, we first rank candidates within a group by score function:
\vspace*{-3mm}
\begin{align}
\textbf{s} = &\> P_\theta(\textbf{y}_t|\textbf{y}_{<t}) + \lambda \cdot \max_{\substack{D(\textbf{a}_\textbf{i}, \textbf{y}) \\ \in \> \textnormal{state S1}}} \frac{|\hat{\textbf{a}}_\textbf{i}| }{|\textbf{a}_\textbf{i}|}
 \end{align}
where $\hat{\textbf{a}}_\textbf{i}$ is $\textbf{a}_\textbf{i}$'s matched prefix with ongoing generation. For example, for $\textbf{y}$ = \emph{``The boy climbs an apple''} and constraint $\textbf{a}_\textbf{i}$=\emph{``apple tree''}, we have $\hat{\textbf{a}}_\textbf{i}$=\emph{``apple''}. The second term denotes maximal percentage of matched prefix in partially satisfied positive literals. Intuitively, this score function ranks candidaite by likelihood and gives a partial reward to candidates moving towards satisfying a positive literal in an unsatisfied clause (state S1). $\lambda$ is an adjustable parameter, controlling how much we favor candidates towards fulfilling another unsatisfied clause. We then proceed in rounds of filling the beam, visiting each group and taking the best scoring ones in rotation, until we reach $k$ candidates. The group traversing order follows the descending order of the highest score in each group.
%
In the end, 
we 
take the highest-scoring hypothesis from the ones with maximal satisfied clauses. 


\begin{table*}[h!]
\footnotesize
    \centering
    \scalebox{0.85}{\begin{tabular}{p{2.2cm}|c||cccccc|c}
     \toprule
    \rowcolor[gray]{0.90}  \textbf{Feature} & \textbf{Example} & CBS & GBS & \citeauthor{post-vilar-2018-fast} & \citeauthor{hu-etal-2019-improved} & CGMH & \citeauthor{sha-2020-gradient} & \textsc{NeuroLogic}\\ 
\midrule
 AND & $oil \land pork$ &$\checkmark$& $\checkmark$& $\checkmark$&$\checkmark$ & $\checkmark$ & $\checkmark$ & $\checkmark$ \\ & Include oil and pork & & & & & & & \\
\midrule
Positive Set AND
 & $oil \land (pork \lor beef)$ & $\checkmark$ & & & & & & $\checkmark$ \\
 & Include oil and a protein & & & & & & & \\
\midrule
Any Predicate 
 & $\neg oil \land (pork \lor beef)$ & & & & &  && $\checkmark$\\
 Logic Formula & Oil-free, include a protein & & & && &  &  \\
\midrule
\midrule
\multicolumn{2}{c}{Runtime:} & $O(Nk2^\mathcal{C})$ & $O(Nk\mathcal{C})$ & $O(Nk)$ & $O(Nk)$  & $O(E)$ & $O(E)$ & $O(Nk)$ \\
    \bottomrule
    \end{tabular}}
    \caption{Expressivity and runtime of various decoding methods. \textit{AND}: Output includes all terms in a set; \textit{Positive Set AND}: Output includes at least one term from each set; \textit{Predicate Logic Formula}: Any combination of positive and negative constraints. $E$ is the number of editing steps, usually much greater than the sequence length $N$. }  %
    \label{tab:constraint_method_table}
\end{table*}

\section{Related Work}
\label{sec:rw}

 \methodshort distinguishes itself from past works in constrained decoding in 3 fundamental ways. 
 \begin{itemize}[topsep=0pt, itemsep=0pt, partopsep=0pt,parsep=0pt]
     \item First, \methodshort generalizes to arbitrary logical constraints by handling the full scope of CNF constraint, while previous works only allow a subset of this (typically conjunctions).
     \item Second, \methodshort effectively optimizes objective function through efficient and diverse search over output space, while previous works suffer from either myopic and narrow or inefficient exploration of the search space.
     \item Third, the asymptotic runtime of \methodshort is $O(Nk)$\footnote{$N$ denotes sequence length and $k$ denotes beam size. In this paper, we the asymptotic runtimes is in terms of the number of calls to a deep generator that scores $P_\theta(\boldsymbol{y}_t|\boldsymbol{y}_{<t})$; this is because calling the generator is the most expensive part of decoding (as opposed to auxiliary bookkeeping).}, same with beam search, constant with respect to number of constraints $\mathcal{C}$. Some previous works suffer from exponential runtime, making  applications infeasible.
 \end{itemize}
 A detailed comparison between \methodshort and previous methods is provided in table \ref{tab:constraint_method_table}.

\subsection{Previous Constrained Decoding Approach}
\citet{anderson-etal-2017-guided} propose constrained beam search (\textbf{CBS}), where constraint satisfaction is tracked by a finite-state machine with $2^{\mathcal{C}}$ states (all possible satisfaction status for $\mathcal{C}$ constraints). Beam search is done over all states with $k$ candidates per state. This method has an exponential complexity $O(Nk2^\mathcal{C})$, making many applications infeasible. 

\citet{hokamp-liu-2017-lexically} propose grid beam search (\textbf{GBS}), which groups together hypotheses by number of constraints satisfied, giving $\mathcal{C}+1$ groups altogether. Each group stores at most $k$ candidates that are expanded at each timestep. GBS has a faster runtime of $O(Nk\mathcal{C})$, but this approach bias towards  sequences satisfying constraints greedily, and collapses into very similar search paths that are often times globally sub-optimal, which results in  dropped language quality.

\citet{post-vilar-2018-fast} propose dynamic beam allocation to reduce GBS's explicit dependence on $\mathcal{C}$. Beam search is done over a single beam, with the $k$ slots of this beam dynamically allocated over the $\mathcal{C}+1$ groups explicitly used by GBS. This approach was made GPU-efficient by \citet{hu-etal-2019-improved}. Still, the language quality issue of GBS remains, and can be worse in practice as fewer hypotheses are considered at each step.

\citet{miao2019cgmh} propose Constrained Generation by Metropolis-Hastings (\textbf{CGMH}). This approach begins by inserting all positive-constraint keywords in random order. Edits are randomly sampled to replace, insert, or delete words to make the sentence fluent; the probability of each action is computed on top of a language model. \citet{sha-2020-gradient} proposes using gradient of a objective function to guide where and how to edit instead of random sampling. These approaches have runtime independent to number of constraints; yet they can involve repeated deletions and insertions, reducing efficiency. Generation quality is also sensitive to initial keyword order and sampled edits.

\subsection{Applications of Constrained Generation}  
Lexically constrained generation can be broadly applied to prior conditional text generation tasks. 
Examples include incorporating pre-specified lexical constraints \cite{anderson-etal-2017-guided, post-vilar-2018-fast},  user-provided terminology constraints \cite{hasler-etal-2018-neural, dinu-etal-2019-training}, noisy automatic constraints \cite{li2019neural} in translation output. A major use case of lexical constrained decoding is paraphrase generation \cite{hu-etal-2019-improved, kajiwara-2019-negative, Hu2019ParaBankMB, miao2019cgmh}, by negatively constraining words in the source to enforce paraphrasing.
Another use case is image captioning, with novel scenes or out-of-domain objects \cite{anderson-etal-2017-guided}, or requiring explicit grounding to objects in the scene \cite{Ren2015FasterRT, Krause2016TheUE}. In addition, \citet{balakrishnan-etal-2019-constrained} leverage constrained decoding to improve semantic correctness for response generation. 


\begin{center}
    \begin{table*}[t]
\setlength{\tabcolsep}{6pt}
\footnotesize
    \centering
    \begin{tabular}{l|c|c c|c|c c|c}
     \toprule
\rowcolor[gray]{0.90} \textbf{\hspace{8pt}Model} &  \multicolumn{1}{c|}{\textbf{ROUGE - L}} &  \multicolumn{2}{c|}{\textbf{BLEU - 3 \& 4}}  & \textbf{METEOR}  &\multicolumn{2}{c|}{\textbf{CIDEr}   \hspace{15pt} \textbf{SPICE}} &\textbf{Coverage}\\ 
\midrule
GPT-2  
&40.3 $\rightarrow$ 42.8&34.2 $\rightarrow$ 36.7&24.7 $\rightarrow$ 26.7&27.6 $\rightarrow$ 30.2&13.4 $\rightarrow$ 14.7 &27.1 $\rightarrow$ 30.3&82.2 $\rightarrow$ 97.7\\
BERT-Gen  
& 42.4 $\rightarrow$ 43.8&37.5 $\rightarrow$ 38.9&27.0 $\rightarrow$ 28.2&29.5 $\rightarrow$ 30.9&14.9 $\rightarrow$ 15.5&29.8 $\rightarrow$ \underline{31.4}&89.2 $\rightarrow$ 97.3\\
UniLM 
&44.3 $\rightarrow$ \textbf{45.8}&40.6 $\rightarrow$ \textbf{42.8}&29.9 $\rightarrow$ \textbf{31.5}&30.1 $\rightarrow$ \textbf{31.7}&15.5 $\rightarrow$ \underline{16.6}&30.6 $\rightarrow$ \textbf{32.5}&90.5 $\rightarrow$ 97.8\\
UniLM-v2  
&43.5 $\rightarrow$ 44.2&39.2 $\rightarrow$ 39.5&28.3 $\rightarrow$ 28.5&30.6 $\rightarrow$ \underline{31.3}&15.2 $\rightarrow$ \textbf{16.8}&30.8 $\rightarrow$ 31.1&92.8 $\rightarrow$ 97.9\\
BART   
&43.3 $\rightarrow$ 44.7&39.9 $\rightarrow$ \underline{41.3}&29.1 $\rightarrow$ \underline{30.6}&30.4 $\rightarrow$ 31.0&15.2 $\rightarrow$ 15.9&30.6 $\rightarrow$ 31.0 &95.0 $\rightarrow$ \textbf{98.7}\\
T5-Large  
&43.9 $\rightarrow$ \underline{44.8}&36.6 $\rightarrow$ 38.5&26.9 $\rightarrow$ 28.1&28.9 $\rightarrow$ 30.7&14.3 $\rightarrow$ 15.5&29.5 $\rightarrow$ 30.8&89.7 $\rightarrow$ \underline{98.5}\\
    \bottomrule
    
    \end{tabular}
    \caption{Experimental results of different supervised models on the C{\scriptsize OMMON}G{\scriptsize EN} test set. Under each column, $\alpha \rightarrow \beta$ shows the performance using the conventional beam search ($\alpha$) compared to the enhanced performance using \textsc{NeuroLogic Decoding} ($\beta$). \textsc{NeuroLogic} always improves the performance across all models and all metrics --- with no exception. The best models are \textbf{bold} and second best ones are \underline{underlined} within each metric.
    }  %
    \label{tab:comGen_result_model}
\end{table*}
\begin{table*}[t]
\footnotesize
    \centering
    \scalebox{.91}{ \begin{tabular}{c|l|c|c c|c|c c|c}
     \toprule

\rowcolor[gray]{0.90} \textbf{Domain Adaption} & \textbf{Model}  &  \multicolumn{1}{c|}{\textbf{ROUGE -  L}} &  \multicolumn{2}{c|}{\textbf{BLEU - 3 \& 4}}  & \textbf{METEOR}  &\multicolumn{2}{c|}{\textbf{CIDEr}   \hspace{5pt} \textbf{SPICE}} &\textbf{Coverage}\\ 
\midrule
  & GPT 
  &26.7 $\rightarrow$ 41.3 &3.0 $\rightarrow$ 25.1 &1.1 $\rightarrow$ 15.9  & \hspace{1.7mm}9.2 $\rightarrow$ 28.8 &0.9 $\rightarrow$ 11.7 &8.0 $\rightarrow$ 29.7& 8.4 $\rightarrow$ \textbf{97.4} \\
No & GPT-2 
&19.7 $\rightarrow$ \textbf{42.9} &4.1 $\rightarrow$ \underline{34.4} &1.5 $\rightarrow$ \underline{23.5} &11.2 $\rightarrow$ \underline{30.7} &0.4 $\rightarrow$ \underline{13.6} &7.1 $\rightarrow$ \underline{31.4} &8.3 $\rightarrow$ 96.0 \\
\midrule
 Yes &GPT-2 
 & 29.8 $\rightarrow$  \underline{42.4} & 9.5 $\rightarrow$ \textbf{36.1} & 4.0 $\rightarrow$ \textbf{25.1}  & 11.7 $\rightarrow$ \textbf{31.3} & 1.7 $\rightarrow$ \textbf{13.9} & 8.0 $\rightarrow$ \textbf{31.8} & 9.3 $\rightarrow$ \underline{96.1} \\
    \bottomrule
    \end{tabular}}
    \caption{Experimental results of different models in zero shot (unsupervised) setting on the C{\scriptsize OMMON}G{\scriptsize EN} test set before and after language domain adaption. Under each column, $\alpha \rightarrow \beta$ shows the performance using the conventional beam search ($\alpha$) compared to the enhanced performance using \textsc{NeuroLogic Decoding} ($\beta$).} %
    \label{tab:comGen_result_zeroshot}
\end{table*}

\end{center}
\newcolumntype{x}[1]{%
>{\centering\hspace{0pt}}p{#1}}%

\begin{table}[t]
\setlength{\tabcolsep}{1pt}
    \centering
    \scalebox{.65}{
    \begin{tabular}{p{2.7cm}| c|x{0.77cm} c|c|x{1.14cm} c|c}
     \toprule

\rowcolor[gray]{0.90}\textbf{Decode Method}  &  \textbf{ROUGE-L} &  \multicolumn{2}{c|}{\textbf{BLEU-3/4}}  & \textbf{METEOR} &\multicolumn{2}{c|}{\textbf{CIDEr}  \textbf{SPICE}} &\textbf{Coverage}\\ 
\midrule
Greedy Decoding &35.3 &25.2 & 16.7&25.8 &10.2 &24.4 & 80.3\\
Top-k Sampling & 33.8& 22.5&14.4 &24.9 &9.2 &22.7 &79.4\\
Top-p Sampling & 35.3& 25.0&16.5 &25.7 &10.2 &24.1 &80.1\\
Beam Search & \underline{40.3} &\underline{34.2} & \underline{24.7}&\underline{27.6} &\underline{13.4} &\underline{27.1} & 82.2\\\midrule
\citeauthor{hokamp-liu-2017-lexically}  & 37.6 & 25.6& 16.8 & 25.9 & 11.1 & 25.1& 97.2\\
\citeauthor{post-vilar-2018-fast} & 38.3&28.1 &18.6 &26.7 &11.8 &26.0 &\underline{97.4} \\
\citeauthor{hu-etal-2019-improved} &38.2 &27.8&18.4 &26.7 &11.7 &26.1 &\underline{97.4} \\
\midrule
\textsc{NeuroLogic} &\textbf{42.8}&\textbf{36.7}&\textbf{26.7}&\textbf{30.2}&\textbf{14.7}&\textbf{30.3}&\textbf{97.7}\\
    \bottomrule
    \end{tabular}}
    \caption{Performance of different decoding methods using supervised GPT2-L on the C{\scriptsize OMMON}G{\scriptsize EN} test set.}  %
    \label{tab:comGen_result_decode}

\end{table}
\vspace*{-9mm}
\section{Experiments I: Constrained Commonsense Generation}
\label{sec:commongen}
C{\small OMMON}G{\small EN} \cite{lin2019commongen} is a benchmark dataset designed as a test of generative commonsense reasoning. Given a set of common concepts (e.g., {dog, frisbee, catch, throw}); the task is to generate a coherent sentence describing an everyday scenario using these concepts (e.g., “a man throws a frisbee and his dog catches it”).

\paragraph{Problem Formulation} The input is an unordered set of $n$ concepts $\textbf{x} = \{a_1, a_2, \ldots , a_n\}$, where each concept $a_i$ is a common object (noun) or action (verb). The expected output is a simple, grammatical sentence $\textbf{y} \in \mathcal{Y}$ that describes a common scenario  using all given concepts in $\textbf{x}$ with correct morphological inflections. 

To apply NeuroLogic Decoding, we impose that each $a_i$ must appear in output $\textbf{y}$ under some morphological inflection. Let $\tilde{a}_i = \{\tilde{a}^i_1, \ldots \tilde{a}^i_{|\tilde{a}_i|}\}$ denote all inflections of $a_i$. $\textbf{y}$ covers concept $a_i$, if at least one of $\{\tilde{a}^i_1, \ldots \tilde{a}^i_{|\tilde{a}_i|}\}$ appears. Formally, 
\vspace*{-3mm}

\[\forall \> a_i \in \textbf{x},\> \exists \> \tilde{a}^i_{j} \in \tilde{a}_i,\> D(\tilde{a}^i_{j}, \textbf{y})\]
where $D(\tilde{a}^i_{j}, \textbf{y})$ is a boolean-value function indicating whether $\textbf{y}$ contains $\tilde{a}^i_j$ or not, as defined above.\footnote{This gets converted into $\land_{i=1}^n \big( \lor_{j=1}^{|\tilde{a}^i|} D(\tilde{a}^i_j, \textbf{y})\big)$.}

\paragraph{Dataset}
The C{\small OMMON}G{\small EN} dataset consists of 35,141 concept-sets (32,651 in \emph{train}, 993 in \emph{val}, 1,497 in \emph{test}) associated with 77,449 sentences. The average size of the concept-sets in the test set is $4.04$, with an average of four sentences per concept-set and an average sentence length of $13.34$ words. 
\paragraph{Approach and Baseline}
The standard pipeline of approaching this problem is to consider it as a conditional sentence generation task. We experiment with several recent pre-trained language models, including GPT-2  \cite{radford2019language}, UniLM \cite{NIPS2019_9464}, UniLM-v2 \cite{bao2020unilmv2}, BERT-Gen \cite{bao2020unilmv2}, BART \cite{lewis-etal-2020-bart}, and T5 \cite{raffel2019exploring}. All models are finetuned with their default hyperparameters. We compare with commonly used decoding methods, including beam search, sampling, and also previously proposed constrained decoding methods. We use several widely-used automatic metrics
to automatically assess the performance, such as BLEU, ROUGE, METEOR, which
mainly focus on measuring surface similarities. We also include metrics specially designed for captioning task, such as CIDEr, and SPICE. Following \citet{lin2019commongen}, we report the concept Coverage, which is the average percentage of input concepts that are present in lemmatizatized outputs.

\subsection{Results I: NeuroLogic vs Other Decoding Methods}
In Table~\ref{tab:comGen_result_decode}, we first present comparisons across different decoding methods based on a supervised sequence-to-sequence model, GPT2. 
The key observations are:

\begin{enumerate}[wide, labelwidth=!,listparindent=0pt, labelindent=0pt,noitemsep,topsep=0pt,parsep=2pt,leftmargin =*]
    \item \textsc{NeuroLogic} outperforms all other previous decoding methods, both constrained and unconstrained, with respect to all metrics and often with a significant margin. 
    \item \textsc{NeuroLogic} not only attains high constraint satisfaction (\textsc{coverage}), it also improves the generation quality as quantified over \textsc{Rouge, Bleu, Meteor, Cide}r, and \textsc{Spice}. 
    \item In comparison, all previous constrained decoding methods \cite{hokamp-liu-2017-lexically, post-vilar-2018-fast, hu-etal-2019-improved} attain high constraint satisfaction at the cost of generation quality; being outperformed here by conventional beam search with a large margin. 
\end{enumerate}

\noindent
The second and the third points above demonstrate that the improved logical expressiveness of \textsc{NeuroLogic} together with the effective search strategy leads to generation that is both higher quality and satisfies the constraints the most effectively. 


\subsection{Results II: NeuroLogic across Different Supervised Models}
Table \ref{tab:comGen_result_model} presents experiments across various state-of-the-art pre-trained language models. In this experiment, all models are supervised on the \textsc{CommonGen} training dataset. 
%
Under each column, $\alpha \rightarrow \beta$ shows the performance using the conventional beam search ($\alpha$) compared to the enhanced performance using \textsc{NeuroLogic Decoding} ($\beta$). 

As before, \textsc{NeuroLogic} always improves the performance across all models and all metrics with no exception -- both in terms of constraint satisfaction as well as generation quality. 
The improvement is especially substantial when the generation quality is relatively low due to smaller model capability or less efficient model architecture or pre-training.

\definecolor{blue1}{HTML}{1F77B4}
\definecolor{purple1}{HTML}{9E0D9E}
\begin{figure}[t]
\centering
    \includegraphics[width=1\linewidth]{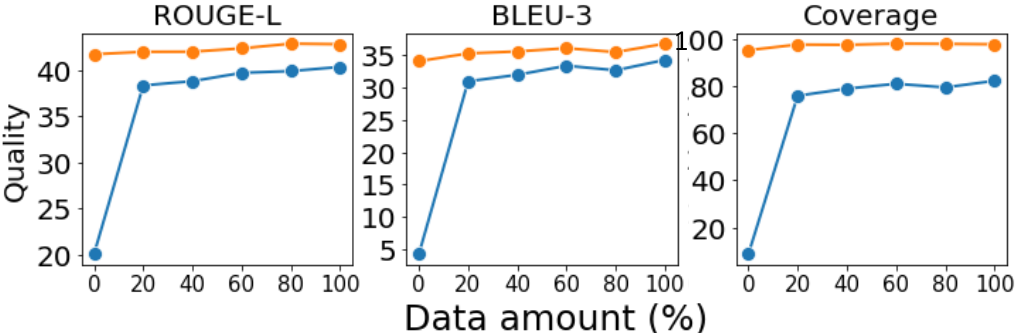}
    \vspace*{-7mm}
     \caption{
     Performance (y-axis) of supervised GPT2-Large on \textsc{CommonGen}, with a varying amount of training data for supervision (x-axis). The \textbf{\textcolor{amber(sae/ece)}{orange line}} denotes decoding with \textsc{NeuroLogic}, and the  \textbf{\textcolor{blue1}{blue line}} denotes decoding with conventional beam search.}
     \label{fig:ablation_training}
     \vspace{4mm}
    \includegraphics[width=0.99\linewidth]{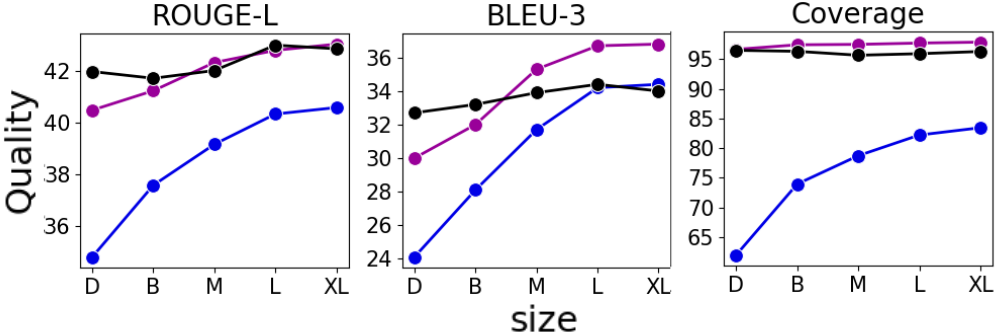}
    \vspace{-7mm}
     \caption{
     Performance (y-axis) of GPT2 with varying model sizes (x-axis). The \textbf{\textcolor{purple1}{purple line}} and \textbf{\textcolor{blue}{blue line}} denote decoding from a supervised model with and without \textsc{NeuroLogic Decoding} respectively. The \textbf{\textcolor{black}{black line}} denotes decoding with \textsc{NeuroLogic} in zero-shot (unsupervised) setting.
     }
     \label{fig:ablation_size}
\end{figure} 

\subsection{Results III: NeuroLogic with Unsupervised Models}
In this experiment, we test how well \textsc{NeuroLogic} works with unsupervised pre-trained language models, with and without domain adaptation. 
%
Table \ref{tab:comGen_result_zeroshot} presents experimental results of zero-shot (i.e., unsupervised) constrained generation. 
With unconstrained decoding, we have zero controllability over the unsupervised language models, as they ignore the problem input and generate irrelevant text. With \textsc{NeuroLogic}, on the other hand, we can dramatically improve the performance on all metrics. 
Fig \ref{tab:comGen_example} demonstrates some generated examples.

In zero-shot setting without any finetuning, the language style of pre-trained LMs might differ from that of C{\small OMMON}G{\small EN}. To further improve the performance, we conduct language domain adaption by fine-tuning the language models on the training-set \textsc{CommonGen} language -- ignoring all concept sets. We observe that after domain adaption, \textsc{NeuroLogic} in zero-shot setting outperforms unconstrained generation with supervised finetuned LMs, which suggests that
inference-time algorithms can provide a more compute-efficient avenue to draw better from neural models. 

\subsection{Results IV: Ablation}
\paragraph{The amount of training data}
\definecolor{blue1}{HTML}{1F77B4}
Figure~\ref{fig:ablation_training} compares the performance (y-axis) of supervised GPT2 with \textsc{NeuroLogic} (\textbf{\textcolor{orange}{ orange line}}) and with conventional beam search (\textbf{\textcolor{blue1}{blue line}}) as a function of the increasing amount of training data (x-axis). Notably, even after being supervised on 100\% of the training data, the supervised GPT2 does not successfully learn the \textsc{CommonGen} constraints (‘coverage’) and is even outperformed by the zero-shot GPT2 (i.e., using 0\% training data) with \textsc{NeuroLogic}.

\paragraph{The model size}
\definecolor{purple1}{HTML}{9E0D9E}
Figure~\ref{fig:ablation_size} compares the 
performance (y-axis) of GPT2 with varying model sizes (x-axis). 
Regardless of the model size, \textsc{NeuroLogic} (\textbf{\textcolor{purple1}{purple line}} and \textbf{\textcolor{black}{black line}}) boosts performance considerably over conventional beam search (\textbf{\textcolor{blue}{blue line}}). More over, if using \textsc{NeuroLogic}, the performance of unsupervised models (\textbf{\textcolor{black}{black line}}) becomes comparable to that of supervised models (\textbf{\textcolor{purple1}{purple line}}). Remarkably, \emph{unsupervised} models with \textsc{NeuroLogic} based on \emph{smaller} networks (\textbf{\textcolor{black}{black line}}) often outperform \emph{supervised} models with conventional beam search based on considerably \emph{larger} networks (\textbf{\textcolor{blue}{blue line}}). 

\newcolumntype{x}[1]{%
>{\centering\hspace{0pt}}p{#1}}%

\begin{table}[t]
\setlength{\tabcolsep}{1pt}
    \centering
    \scalebox{.71}{
    \begin{tabular}{p{2.7cm}|c|x{0.77cm} c|c|c|c}
     \toprule

\rowcolor[gray]{0.90}\textbf{Decode Method}  &  \textbf{ROUGE-L} &  \multicolumn{2}{c|}{\textbf{BLEU-3/4}}  & \textbf{METEOR} &\textbf{Coverage} &\textbf{Extra}\\ 
\midrule
Top-k Sampling  & 27.5& 15.2&9.5 &19.2  &84.8 &16.0\\
Top-p Sampling &28.7 &\underline{17.6} &11.7 &19.4  &86.4 &15.4\\
Beam Search &\underline{29.4} &17.4 & \underline{12.0}&\underline{19.7}  &86.5 &14.3\\\midrule
\citeauthor{post-vilar-2018-fast} & 26.1&13.6 &8.8 &16.5  &\underline{89.6} &1.15 \\
\citeauthor{hu-etal-2019-improved} &26.1 &13.6 &8.8 &16.5  &\underline{89.6} &\underline{1.13} \\
\midrule
\textsc{NeuroLogic} &\textbf{32.1}&\textbf{19.5}&\textbf{13.8}&\textbf{19.8}&\textbf{95.8}&\textbf{0.6}\\
    \bottomrule
    \end{tabular}}
     \caption{Experimental results of different decoding methods with RecipeGPT on the Recipe1M+ test set. \emph{Coverage} indicates the average percentage of ingredients that are covered in the generated recipe, while \emph{Extra} corresponds to the average ratio of hallucinated ingredients over the number of given ingredients.} 
    \label{tab:result_recipe}

\end{table}
\section{Experiments II: Recipe Generation}
We next study cooking recipe generation, a paragraph-level generation task. Given a dish name and a list of ingredients, the task is to generate cooking instructions for the given recipe. 


\paragraph{Problem Formulation}  The input is the recipe title, an unordered set of ingredients $E = \{e_1, ..., e_{|E|}\}$ where $e_i$ can be a single- or multi-word ingredient phrase (e.g., `onions', `black pepper'). Let $\mathcal{G}$ denote the set of all ingredients. The expected output is a paragraph $\textbf{y} \in \mathcal{Y}$ that describes multi-step cooking instructions.  

To apply \textsc{NeuroLogic Decoding}, we constrain output $\textbf{y}$ to contain all given ingredients $e_i$ in $E$, and no other ingredients, i.e. no ingredients in $\mathcal{G} \setminus E$. Ingredients can be referred to with generic terms (e.g., `vegetables' may refer to `onions', or `carrots') and we denote the generic name for ingredient $e_i$ as $e^T_i$. Formally, the constraint is
\vspace*{-5mm}

\begin{small}
\begin{equation*}\begin{split}
&\Big(\forall e_i \in E, D(e_i, \textbf{y}) \lor D(e^T_i, \textbf{y})\Big) \\&\land \Big(\forall e_i \in \mathcal{G} \setminus E, \neg D(e_i, \textbf{y})\Big )
\end{split}
\end{equation*}
\end{small}

\vspace*{-5mm}


\paragraph{Dataset, Approach and Baseline} 
We use Recipe1M+, a large-scale, structured corpus of over one million cooking recipes. On average each recipe has 118 words and 9 ingredients.
RecipeGPT \cite{Lee2020RecipeGPTGP} is a GPT-2 model fine-tuned on Recipe1M+, for generating recipes. Its default decoding algorithms are beam search and sampling, which serve as the baselines for evaluating our method. In addition, we compare against previously proposed constrained decoding methods with RecipeGPT.
Besides common evaluation metrics for generation task, we introduce explicit measures of given-ingredient coverage and usage of extra/hallucinated ingredients.

\paragraph{Result} Table \ref{tab:result_recipe} presents the experimental results. We can see that \textsc{NeuroLogic} outperforms all baselines in all metrics.  The delta is quite remarkable on coverage of given ingredients and usage of extra ingredients. With \textsc{NeuroLogic}, we are able to cover almost all ingredients in generated instructions and guarantee not to use any other ingredients, which leads to more accurately controlled generation. By plugging \textsc{NeuroLogic} into existing generation system, we can get immediate boosts in controllability and generation quality with no extra computational cost. 

\newcolumntype{x}[1]{%
>{\centering\hspace{0pt}}p{#1}}%

\begin{table}[t]
\footnotesize
    \centering
     \scalebox{.93}{
    \begin{tabular}{@{}p{4.1em}|p{5.0em}@{\hspace{0.1em}}c@{\hspace{1.1em}}c@{\hspace{1.1em}}c@{}}
     \toprule
\rowcolor[gray]{0.93} Supervised? & \textbf{Model}&  \textbf{ROUGE-L}&  \textbf{BLEU-4} & \textbf{METEOR} \\ 
\midrule
Yes & GPT2  & 70.5 | \textbf{72.6} & 87.6 | \textbf{92.4} & 60.0 | \textbf{64.0} \\
Yes & BART & \underline{72.9} | 70.2& 89.5 | 87.0 & 60.2 | 54.2\\
Yes & T5  & 70.9 | 69.9 & 82.4 | 79.7 &54.6 | 50.4\\
Yes & \citeauthor{kiddon-etal-2016-globally} & -  & \underline{90.6} | 77.8 & \underline{62.1} | 54.4\\
\midrule
\textbf{No} & { GPT2 +}  & \textbf{73.9} | \underline{71.8} & \textbf{94.8} | \underline{90.8} & \textbf{66.6} | \underline{62.0} \\
 & {\scriptsize \textsc{NeuroLogic}}  &  & &  \\
    \bottomrule
    
    \end{tabular}}
    \caption{Results of dialogue generation, the right column is generation result for hotel systems, the left column is for restaurant systems}  %
    \label{tab:diag}
\end{table}
\section{Experiments III: Data-Grounded Dialogue Response Generation}

In dialogue responses generation for hotel and restaurant information systems \cite{wen-etal-2016-multi}, we generate a natural language response given a query type (e.g., informing or querying) and a list of facts to convey (e.g., a hotel’s name and address). 

\paragraph{Problem Formulation} 
The input is query type, unordered set of facts $F = \{f_1, ..., f_{|F|}\}$, where each $f_i$ contains attribute and value (i.e. accepts\_credit\_cards=``yes'', name=``red victorian bed breakfast''). The expected output is a dialogue responses $\textbf{y} \in \mathcal{Y}$ containing given information. 

The lexical constraint here is that all given facts $f_i$ must be included in responses $\textbf{y}$ in proper natural language form $f^N_i$. We use a very simple template to turn $f_i$ to natural language form $f^N_i$.  (i.e. the natural language form for accepts\_credit\_cards=``no'' is ``doesn't accept credit cards''). Formally,
\vspace*{-1mm}
\[\forall \> f_i \in F,\> D(f^N_i, \textbf{y})\]
\vspace*{-8mm}
\paragraph{Dataset, Approach and Baseline}
We use the hotel and restaurant dialogue system corpus and the same train-dev-test split from \cite{wen-etal-2016-multi}. There are 8 query types and 12 attribute types.

The standard paradigm for dialogue generation is to consider it as a conditional  sentence generation task and finetune a seq2seq model. While this pipeline works effectively with existing data, but once we have user queries with new query types or new attribute types, the seq2seq model would not be able to generate plausible responses. The situation can happen frequently with a dialogue generation system in application. Thus, we are interested in zero-shot dialogue generation. We give a hand-crafted initial prompt to a pre-trained LM based on the query type and apply \textsc{NeuroLogic Decoding} to force given facts to include in generation. The pre-trained LM we use here is GPT2 \cite{radford2019language}.

The baseline we compare against is seq2seq finetuned LMs with vanilla beam search, including GPT-2 \cite{radford2019language}, BART \cite{lewis-etal-2020-bart} and T5 \cite{raffel2019exploring}. We also compare with previous SOTA \cite{kiddon-etal-2016-globally} on dialogue response generation. 

\paragraph{Result}
Table \ref{tab:diag} presents the experimental results. We can see that zero-shot generation with proposed method outperforms or matches supervised baselines. This suggests that plugging \textsc{NeuroLogic Decoding} into pretrained LMs can lead to a powerful dialogue generation system, we do not actually need massive finetuning with extra computational cost to do that.
%

\begin{table}[t]
\footnotesize
    \centering
        \begin{tabular}{@{}l@{\hspace{0.1em}}l@{\hspace{0.0em}} c@{\hspace{0.4em}}c@{}}
        \toprule
\rowcolor[gray]{0.95} & \textbf{Model} & \textbf{Accuracy}(\%; $\uparrow$) & \textbf{$\Delta_{S}$} (F1; $\downarrow$) \\ \midrule
\multirow{4}{*}{\rotatebox[origin=c]{90}{En-De}} & Google Translate & 59.4 \phantom{$\rightarrow$ \textbf{81.0}}& 12.5\phantom{$\rightarrow$ \textbf{4.3}}\\
& Microsoft Translator & 74.1\phantom{$\rightarrow$ \textbf{91.0}} & 30.2\phantom{$\rightarrow$ \textbf{4.3}}\\
& {\scriptsize \citeauthor{mariannmt}} & 60.5 $\rightarrow$ \textbf{91.0} & 13.3 $\rightarrow$ \textbf{4.3}\\
& {\scriptsize \citeauthor{mariannmt}+GT Gender} & 60.5 $\rightarrow$ \textbf{95.0} & 13.3 $\rightarrow$ \textbf{2.4}\\ 
\midrule
\multirow{4}{*}{\rotatebox[origin=c]{90}{En-Fr}} & Google Translate & 63.6 \phantom{$\rightarrow$ \textbf{81.0}}&26.7\phantom{$\rightarrow$ \textbf{4.3}}\\
& Microsoft Translator & 44.7 \phantom{$\rightarrow$ \textbf{81.0}}&29.7\phantom{$\rightarrow$ \textbf{4.3}}\\
& {\scriptsize \citeauthor{mariannmt}} & 53.0 $\rightarrow$ \textbf{81.0} & 19.3 $\rightarrow$ \textbf{1.7}\\
& {\scriptsize \citeauthor{mariannmt} +GT Gender} & 53.0 $\rightarrow$ \textbf{89.9} & 19.3 $\rightarrow$ \textbf{1.5}\\
    \bottomrule
    \end{tabular}
    
    \caption{Performance of Gender Bias Removal on WinoMT, adapted from \citeauthor{stanovsky-etal-2019-evaluating}.  Accuracy refers to correctly translating a person's gender, $\Delta_{S}$ is the difference in performance ($F_1$) between stereotypical and non-stereotypical gender roles (lower is better). The arrows ($\rightarrow$) show the results before and after \textsc{NeuroLogic Decoding}, where gender is inferred from a coreference model (default) or provided (GT Gender).}  %
    \label{tab:result_gender}
\end{table}

\section{Experiment IV: Reducing Gender Bias in Machine Translation}

\paragraph{Problem Formulation} 
We adopt the task setup and dataset of \citet{stanovsky-etal-2019-evaluating}. The input $x$ is an English sentence describing a scenario with human entities $N = \{n_1, \ldots, n_{|N|}\}$ who are identified by roles. The desired output is a translation $\textbf{y}$ which uses the correct gender inflections in the target language (here, German or French).

We obtain indicators of people's gender identity through coreference resolution, linking each entity with their gendered pronoun.\footnote{We could use any off-the-shelf coreference resolution model for this. However, since the English examples in \citet{stanovsky-etal-2019-evaluating} follow the Winograd schemas format, we use a RoBERTa model finetuned on Winograd Schema Challenge for this, with 78.4\% accuracy.}
We then constrain the correctly-gendered human entities to appear in output $\textbf{y}$. For a human entity $n_i$, let $n^F_i$ denote its female inflection in the target language, and $n^M_i$ denotes its male inflection. Let $F$ denotes the set of human entities associated with female characters, and $M$ denotes the set of entities associated with male. Formally, the constraint is 
\vspace*{-3mm}

\begin{small}
\begin{equation*}\begin{split}
    \Big(\forall n_i \in F,D(n^F_i, \textbf{y}) \land \neg D(n^M_i, \textbf{y})\Big) \land \\
    \Big(\forall n_i \in M,D(n^M_i, \textbf{y}) \land \neg D(n^F_i, \textbf{y}) \Big) 
\end{split}
\end{equation*}

\end{small}

\paragraph{Dataset} We use \citet{stanovsky-etal-2019-evaluating}'s dataset, which is built over the English-only coreference gender-bias studies: Winogender \cite{rudinger2018gender} and Wino-Bias \cite{zhao2018gender}.

\paragraph{Result} 
Our results are shown in Table~\ref{tab:result_gender}. When provided gender markers given by a coreference model, \textsc{NeuroLogic Decoding} increases the accuracy of handling gender correctly by \textbf{30.5} percentage for German, and \textbf{28.0} percentage for French. This even outperforms commercial translation systems -- the best result, over any language or system, is Microsoft Translator for German with 74.1\% accuracy, whereas \textsc{NeuroLogic Decoding} enables the baseline model to get 91\% accuracy. The performance increases again by an additional 4\% (German) and 8.9\% (French) when ground-truth gender markers are used during constrained decoding. Last, the diagnostic results also show that \textsc{NeuroLogic Decoding} is particularly effective at reducing (over)reliance on stereotypical gender roles, with a significant decrease in performance difference $\Delta_S$ between stereotypical and non-stereotypical gender roles. These results suggest that \textsc{NeuroLogic Decoding} a plug-and-play approach for reducing gender bias in existing translation systems. 
%

\section{Conclusion}

We propose \textsc{NeuroLogic Decoding}, an efficient and general method for generating with arbitrary predicate logic constraints. We demonstrate its intuitive application to $4$ different tasks as an extension to existing models, showing broad and consistent improvement to decoding quality. 

\section*{Acknowledgements}

We thank the anonymous reviewers and meta-reviewers for their helpful feedback.
This research was supported in part by DARPA under the MCS program through NIWC Pacific (N66001-19-2-4031) and the Allen Institute for AI (AI2).

\bibliography{anthology,custom}
\bibliographystyle{acl_natbib}

\clearpage

\appendix


\begin{figure*}[t]
\vspace{-1mm}
\centering
    \includegraphics[width=0.87\linewidth]{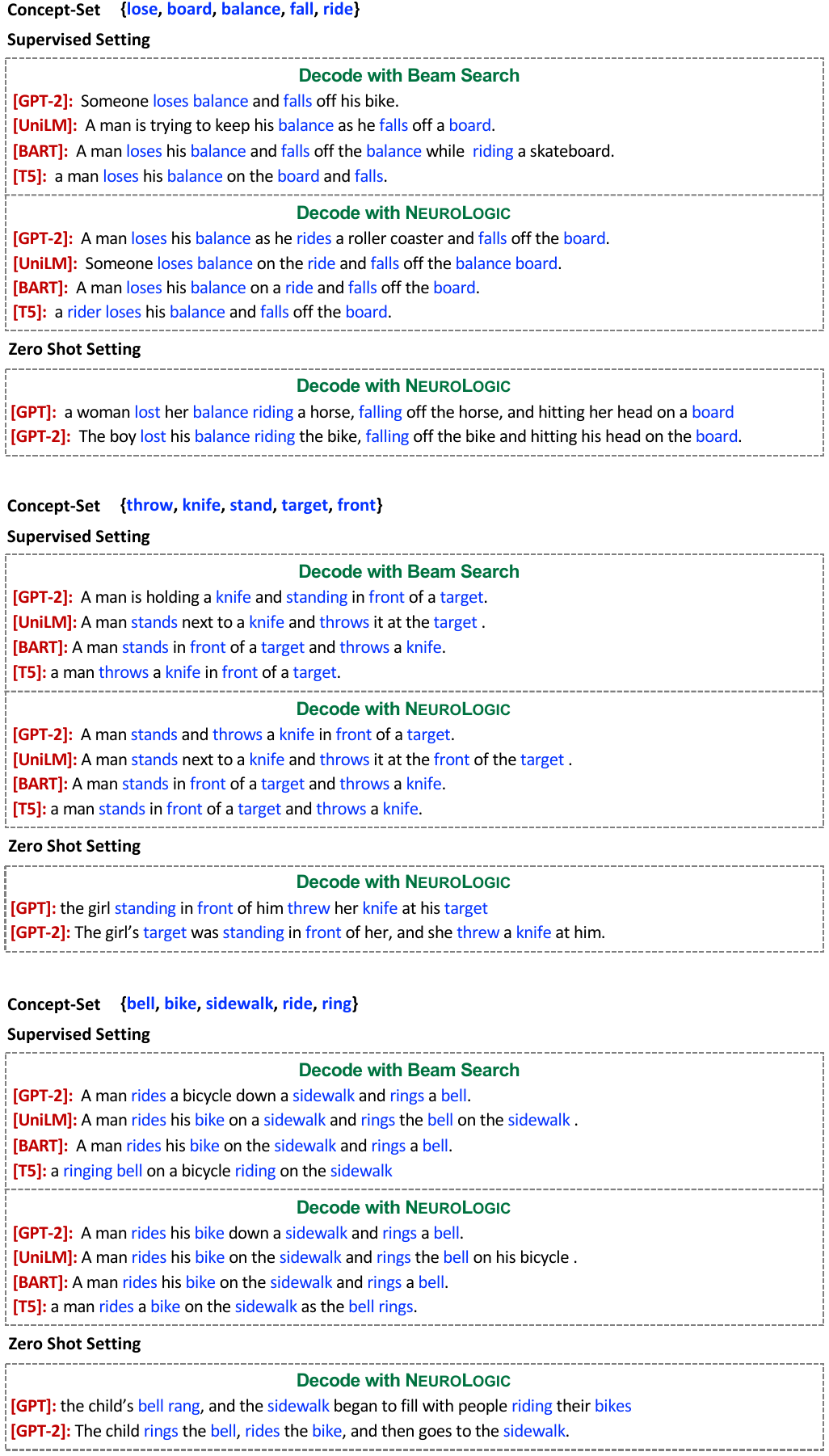}
    \vspace{-1mm}
     \caption{Generation examples of different models in supervised and zero-shot setting with and without \textsc{NeuroLogic Decoding}, on \textsc{\commongen.}} 
     \label{tab:comGen_example}
\end{figure*} 
\end{document}